# PARSING TURKISH USING THE LEXICAL FUNCTIONAL GRAMMAR FORMALISM[1]

**Zelal Güngördü**

Centre for Cognitive Science
University of Edinburgh
Edinburgh, Scotland, U.K.
gungordu@cogsci.ed.ac.uk

**Kemal Oflazer**

Department of Computer Engineering
Bilkent University
Bilkent, Ankara, Turkey
ko@cs.bilkent.edu.tr

**Abstract** This paper describes our work on parsing Turkish using the *lexical-functional grammar* formalism. This work represents the first effort for parsing Turkish. Our implementation is based on Tomita's parser developed at Carnegie-Mellon University Center for Machine Translation. The grammar covers a substantial subset of Turkish including simple and complex sentences, and deals with a reasonable amount of word order freeness. The complex agglutinative morphology of Turkish lexical structures is handled using a separate two-level morphological analyzer. After a discussion of key relevant issues regarding Turkish grammar, we discuss aspects of our system and present results from our implementation. Our initial results suggest that our system can parse about 82% of the sentences directly and almost all the remaining with very minor pre-editing.

## 1 INTRODUCTION

As part of our ongoing work on the development of computational resources for natural language processing in Turkish we have undertaken the development of a parser for Turkish using the lexical-functional grammar formalism, for use in a number of applications. This work represents the first approach to the computational analysis of Turkish, though there have been a number of studies of Turkish syntax from a linguistic perspective (e.g., [Meskill 1970]). Our implementation is based on Tomita's parser developed at Carnegie-Mellon University Center for Machine Translation [Musha et.al. 1988, Tomita 1987]. Our grammar covers a substantial subset of Turkish including simple and complex sentences, and deals with a reasonable amount of word order freeness.

Turkish has two characteristics that have to be taken into account: agglutinative morphology, and rather free word order with explicit case marking. We handle the rather complex agglutinative morphology of the Turkish lexical structures using a separate morphological processor based on the two-level paradigm [Antworth 1990, Oflazer 1993] that we have integrated with the lexical-functional grammar parser. Word order freeness is dealt with by relaxing the order of phrases in the phrase structure parts of lexical-functional grammar rule by means of generalized phrases.

## 2 LEXICAL-FUNCTIONAL GRAMMAR

Lexical-functional grammar (LFG) is a linguistic theory which fits nicely into computational approaches that use *unification* [Shieber 1986]. A lexical-functional grammar assigns two levels of syntactic description to every sentence of a language: a *constituent structure* and a *functional structure*. Constituent structures (c-structures) characterize the phrase structure configurations as a conventional phrase structure tree, while surface grammatical functions such as *subject, object*, and *adjuncts* are represented in functional structure (f-structure). Because of space limitations we will not go into the details of the theory. One can refer to Kaplan and Bresnan [Kaplan and Bresnan 1982] for a thorough discussion of the LFG formalism.

## 3 TURKISH SYNTAX

In this section, we would like to highlight two of the relevant key issues in Turkish grammar, namely highly inflected agglutinative morphology and free word order, and give a description of the structural classification of Turkish sentences that we deal with.

### 3.1 Morphology

Turkish is an agglutinative language with word structures formed by productive affixations of derivational and inflectional suffixes to root words [Oflazer 1993]. This extensive use of suffixes causes morphological parsing of words to be rather complicated, and results in ambiguous lexical interpretations in many cases. For example:

(1) **çocukları**
a. child+PLU+3SG-POSS    his children
b. child+3PL-POSS        their child
c. child+PLU+3PL-POSS    their children
d. child+PLU+ACC         children (acc.)

Such ambiguity can sometimes be resolved at phrase and sentence levels by the help of agreement requirements though this is not always possible:

(2a) **Onların**      **çocukları**       **geldiler.**
     it+PLU+GEN       child+PLU           come+PAST
                     +3PL-POSS           +3PL
     (Their          children             came.)

---

Table 1: Percentage of different word orders in Turkish.

| Sentence Type | Children Speech | Adult Speech |
|---|---|---|
| SOV | 46% | 48% |
| OSV | 7% | 8% |
| SVO | 17% | 25% |
| OVS | 20% | 13% |
| VSO | 10% | 6% |
| VOS | 0% | 0% |

(2b) **Çocukları geldiler.**
    child+PLU+3SG-POSS come+PAST+3PL
    (His children came.)
    child+PLU+3PL-POSS come+PAST+3PL
    (Their children came.)

For example, in (2a) only the interpretation (1c) (i.e., *their children*) is possible because:

- the agreement requirement between the modifier and the modified parts in a possessive compound noun eliminates (1a).[2]

- the facts that *gel (come)* does not subcategorize for an accusative marked direct object, and that in Turkish the subject of a sentence must be nominative[3] eliminate (1d).

- the agreement requirement between the subject and the verb of a sentence eliminates (1b).[4]

In (2b), both (1a) and (1c) are possible (*his children*, and *their children*, respectively) because the modifier of the possessive compound noun is a covert one: it may be either *onun* (*his*) or *onların* (*their*). The other two interpretations are eliminated due to the same reasons as in (2a).

### 3.2 Word Order

If we concern ourselves with the *typical order of constituents*, Turkish can be characterized as being a *subject–object–verb (SOV) language*, though the data in Table 1 from Erguvanlı [Erguvanlı 1979], shows that other orders for constituents are also common (especially in discourse). In Turkish it is not the position, but the case of a noun phrase that determines its grammatical function in the sentence. Consequently typical order of the constituents may change rather freely without affecting the grammaticality of a sentence. Due to various syntactic and pragmatic constraints, sentences with the non-typical orders are not

---

[2]The agreement of the modifier must be the same as the possessive of the modified with the exception that if the modifier is third person plural the possessive of the modified may be third person singular.

[3]In Turkish, the nominative case is unmarked.

[4]In a Turkish sentence, person features of the subject and the verb should be the same. This is true also for the number features with one exception: third person plural subjects may sometimes take third person singular verbs.

stylistic variants of the typical versions which can be used interchangeably in any context [Erguvanlı 1979]. For example, a constituent that is to be emphasized is generally placed immediately before the verb. This affects the places of all the constituents in a sentence except that of the verb:

(3a) **Ben çocuğa kitabı verdim.**
    I    child+DAT book+ACC give+PAST+1SG
    (I gave the book to the child.)

(3b) **Çocuğa kitabı <u>ben</u> verdim.**
    child+DAT book+ACC I give+PAST+1SG
    (It was me who gave the child the book.)

(3c) **Ben kitabı <u>çocuğa</u> verdim.**
    I book+ACC child+DAT give+PAST+1SG
    (It was the child to whom I gave the book.)

(3a) is an example of the typical word order whereas in (3b) the subject, *ben*, is emphasized. Similarly, in (3c) the indirect object, *çocuğa*, is emphasized.

In addition to these possible changes, the verb itself may move away from its typical place, i.e., the end of the sentence. Such sentences are called *inverted sentences* and are typically used in informal prose and discourse.

However, this looseness of ordering constraints at sentence level does not extend into all syntactic levels. There are even constraints at sentence level:

• A nominative direct object should be placed immediately before the verb.[5] Hence, (5b) is ungrammatical:[6]

(5a) **Ben çocuğa kitap verdim**.
    I child+DAT book give+PAST+1SG
    (I gave a book to the child.)

(5b) *****Çocuğa kitap ben verdim.**
    child+DAT book I give+PAST+1SG

• Some adverbial complements of quality (those that are actually qualitative adjectives) always precede the verb or, if it exists, the indefinite direct object:

(6a) **Yemeği iyi pişirdin.**
    meal+ACC good cook+PAST+2SG
    (You cooked the meal well.)

(6b) **İyi yemeği pişirdin.**
    good meal+ACC cook+PAST+2SG
    (You cooked the good meal.)

(6c) **İyi yemek pişirdin.**
    good meal cook+PAST+2SG
    (You cooked a good meal./You cooked a meal well.)

Note that although (6b) is grammatical *iyi* is no more an adverbial complement, but is an adjective that modifies *yemeği*. Note also that (6c) is ambiguous: *iyi* can be interpreted either as an adjective modifying *yemek* or as an

---

[5]In Turkish, a transitive verb that subcategorizes for a direct object, can take either an accusative marked or a nominative marked (unmarked on the surface) noun phrase for that object. The function of accusative case marking is to indicate that the object refers to a particular definite entity, though there are very rare cases where this is not the case.

[6]Note that (3b,c) are grammatical since the direct object, *kitabı*, is marked accusative.

adverb modifying *pişirdin*.[7]

## 3.3 Structural Classification of Sentences

The following summarizes the major classes of sentences in Turkish.

•**Simple Sentences:** A simple sentence contains only one independent judgement. The sentences in (2), (3), (4a), (5a), and (6) are all examples of simple sentences.

•**Complex Sentences :** In Turkish, a sentence can be transformed into a construction with a *verbal noun*, a *participle* or a *gerund* by affixing certain suffixes to the verb of the sentence. Complex sentences are those that include such dependent (subordinate) clauses as their constituents, or as modifiers of their constituents. Dependent clauses may themselves contain other dependent clauses. So, we may have embedded structures such as:

(7) **Burada** **içilebilecek** **su**
here+LOC drink+PASS+POT water
+FUT+PART
**bulamayacağımı** **zannetmek** **doğru**
find+NEG-POT think+INF right
+FUT+PART
+1SG-POSS
+ACC
**olmazdı.**
be+NEG+AOR
+PAST+3SG

(It wouldn't have been right for me to think that I wouldn't be able to find drinkable water here.)

The subject of (7) (*burada içilebilecek su bulamayacağımı zannetmek - to think that I wouldn't be able to find drinkable water here*) is a nominal dependent clause whose definite object (*burada içilebilecek su bulamayacağımı – that I wouldn't be able to find drinkable water here*) is an adjectival dependent clause which acts as a nominal one. The indefinite object of this definite object (*içilebilecek su – drinkable water*) is a compound noun whose modifier part is another adjectival dependent clause (*içilebilecek – drinkable*), and modified part is a noun (*su – water*).

It should be noted that there are other types of sentences in the classification according to structure. However, we will not be concerned with them here because of space limitations. (See Şimşek [Şimsek 1987], and Güngördü [Güngördü 1993] for details.)

## 4 SYSTEM ARCHITECTURE AND IMPLEMENTATION

We have implemented our parser in the grammar development environment of the Generalized LR Parser/Compiler developed at Carnegie Mellon University Center for Machine Translation. No attempt has been made to include morphological rules as the parser lets us incorporate our own morphological analyzer for which we use a full scale two-level specification of Turkish morphology based on a lexicon of about 24,000 root words[Oflazer 1993]. This lexicon is mainly used for morphological analysis and has limited additional syntactic and semantic information, and is augmented with an argument structure database.[8]

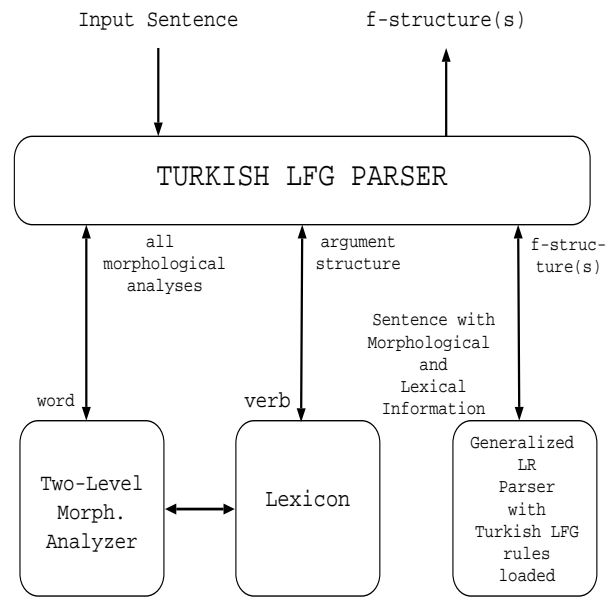

Figure 1: The system architecture.

Figure 1 shows the architecture of our system. When a sentence is given as input to the program, the program first calls the morphological analyzer for each word in the sentence, and keeps the results of these calls in a list to be used later by the parser.[9] If the morphological analyzer fails to return a structure for a word for any reason (e.g., the lexicon may lack the word or the word may be misspelled), the program returns with an error message. After the morphological analysis is completed, the parser is invoked to check whether the sentence is grammatical. The parser performs bottom-up parsing. During this analysis, whenever it consumes a new word from the sentence, it picks up the morphological structure of this word from the list. If the word is a finite verb or an infinitival, the parser is also provided with the subcategorization frame of the word. At the end of the analysis, if the sentence is grammatical, its f-structure is output by the parser.

---

[7]The second interpretation is possible since *yemek* is an indefinite direct object.

[8]The morphological analyzer returns a list of *feature–value* pairs. For instance for the word *evdekilerin* (of those (things) in the house/your things in the house) it will return

1. ((*CAT* N)(*R* "ev")(*CASE* LOC)(*CONV* ADJ "ki")(*AGR* 3PL)(*CASE* GEN))

2. ((*CAT* N)(*R* "ev")(*CASE* LOC)(*CONV* ADJ "ki")(*AGR* 3PL)(*POSS* 2SG))

[9]Recall that there may be a number of morphologically ambiguous interpretations of a word. In such a case, the morphological analyzer returns all of the possible morphological structures in a list, and the parser takes care of the ambiguity regarding the grammar rules.

Table 2: The number of rules for each category in the grammar.

| Category | Number of Rules |
| --- | --- |
| Noun phrases | 17 |
| Adjectival phrases | 10 |
| Postpositional phrases | 24 |
| Adverbial constructs | 50 |
| Verb phrases | 21 |
| Dependent clauses | 14 |
| Sentences | 6 |
| Lexical look up rules | 11 |
| TOTAL | 153 |

## 5  THE GRAMMAR

In this section, we present an overview of the LFG specification that we have developed for Turkish syntax. Our grammar includes rules for *sentences, dependent clauses, noun phrases, adjectival phrases, postpositional phrases, adverbial constructs, verb phrases*, and a number of *lexical look up rules*.[10] Table 2 presents the number of rules for each category in the grammar. There are also some intermediary rules, not shown here.

Recall that the typical order of constituents in a sentence may change due to a number of reasons. Since the order of phrases is fixed in the phrase structure component of an LFG rule, this rather free nature of word order in sentence level constitutes a major problem. In order to keep from using a number of redundant rules we adopt the following strategy in our rules: We use the same place holder, $<XP>$, for all the syntactic categories in the phrase structure component of a sentence or a dependent clause rule, and check the categories of these phrases in the equations part of the rule. In Figure 2, we give a grammar rule for the sentence with two constituents, with an informal description of the equation part.[11]

Recall also that an indefinite object should be placed immediately before the verb, and some adverbial complements of quality (those that are actually qualitative adjectives) always precede the verb or, if it exists, the indefinite direct object. In our grammar, we treat such objects and adverbial complements as parts of the verb phrase. So, we do not check these constraints at the sentence or dependent clause level.

## 6  PERFORMANCE EVALUATION

In this section, we present some results about the performance of our system on test runs with four different texts on different topics. All of the texts are articles taken from magazines. We used the CMU Common Lisp system running

---

[10] Recall that no morphological rules are included. The lexical look up rules are used just to call the morphological analyzer.

[11] Note that *x0, x1*, and *x2* refer to the functional structures of the sentence, the first constituent and the second constituent in the phrase structure, respectively.

```
(<S> <==> (<XP> <XP>)
  1) if x1's category is VP then
       assign x1 to the functional structure
       of the verb of the sentence
     if x2's category is VP then
       assign x2 to the functional structure
       of the verb of the sentence

  2) for i = 1 to 2 do

       if xi has already been assigned to
        the verb then do nothing

       if xi's category is ADVP then
        add xi to the adverbial complements
        of the sentence

       if xi's category is NP and
        xi's case is nominative then
        assign xi to the functional struct-
        ure of the subject of the sentence

       if xi's category is NP then
        if the verb of the sentence can take
         an object with this case (consider
         also the voice of the verb)
         add xi to the objects of the verb

  3) check if the verb has taken all the
     objects that it has to take

  4) make sure that the verb has not
     taken more than one object with
     the same  thematic role

  5) check if the subject and the verb
     agree in  number and person:
      if the subject is defined (overt)
      then
        if the agreement feature of the
          subject is third person plural
        then the agreement feature of the
          verb may be either third person
          singular or third person plural
        else
          the agreement features of the
          subject and the verb must be
          the same
      else if the subject is undefined
          (covert) then assign the
          agreement feature of the verb
          to that of the subject
```

Figure 2: An LFG rule for the sentence level given with an informal description of the equation part.

Table 3: Statistical information about the test runs.

| Doc | #S | #S in Scope | #S ign. | #S after Pre-ed. | #P per Sent. | Secs per Sent. |
|---|---|---|---|---|---|---|
| 1 | 43 | 30 | 0 | 55 | 4.28 | 12.26 |
| 2 | 51 | 41 | 2 | 62 | 5.02 | 8.92 |
| 3 | 56 | 48 | 1 | 64 | 4.87 | 10.28 |
| 4 | 80 | 70 | 0 | 97 | 3.25 | 7.46 |
| Tot. | 230 | 189 | 3 | 279 | – | – |
|  | 100% | 82% |  |  |  |  |

#S: Number of sentences, #P: Number of parses.

in a Unix environment, on SUN Sparcstations at Center for Cognitive Science, University of Edinburgh.[12]

In all of the texts there were some sentences outside our scope. These were:

- sentences that contain finite sentences as their constituents or modifiers of their constituents,
- conditional sentences,
- finite sentences that are connected by coordinators (and/or), and
- sentences with discontinuous constituents.[13]

We pre-edited the texts so that the sentences were in our scope (e.g., separated finite sentences connected by coordinators and parsed them as independent sentences, and ignored the conditional sentences). Table 3 presents some statistical information about the test runs. The first, second and third columns show the document number, the total number of sentences and the number of sentences that we could parse without pre-editing, respectively. The other columns show the number of sentences that we totally ignored, the number of sentences in the pre-edited versions of the documents, average number of parses per sentence generated and average CPU time for each of the sentences in the texts, respectively. It can be seen that our grammar can successfully deal with about 82% of the sentences that we have experimented with, with almost all remaining sentences becoming parsable after a minor pre-editing. This indicates that our grammar coverage is reasonably satisfactory.

Below, we present the output for a sentence which shows very nicely where the structural ambiguity comes out in Turkish.[14] The output for (8a) indicates that there are four ambiguous interpretations for this sentence as indicated in (8b-e):[15]

(8a) **Küçük   kırmızı   top   gittikçe   hızlandı.**
little   red   ball   go+GER   speed up
   red paint+   gradually   +PAST
   3SG–POSS       +3SG

(8b) The little red ball gradually sped up.
(8c) The little red (one) sped up as the ball went.
(8d) The little (one) sped up as the red ball went.
(8e) It sped up as the little red ball went.

The output of the parser for the first interpretation is given in Figure 3. This output indicates that the subject of the sentence is a noun phrase whose modifier part is *küçük*, and modified part is another noun phrase whose modifier part is *kırmızı* and modified part is *top*. The agreement of the subject is third person singular, case is nominative, etc. *Hızlandı* is the verb of the sentence, and its voice is active, tense is past, agreement is third person singular, etc. *Gittikçe* is a temporal adverbial complement.

Figures 4 through 7 illustrate the c-structures of the four ambiguous interpretations (8b-e), respectively:[16]

- In (8b), the adjective *kırmızı* modifies the noun *top*, and this noun phrase is then modified by the adjective *küçük*. The entire noun phrase functions as the subject of the main verb *hızlandı*, and the gerund *gittikçe* functions as an adverbial adjunct of the main verb.

- In (8c), the adjective *kırmızı* is used as a noun, and is modified by the adjective *küçük*.[17] This noun phrase functions as the subject of the main verb. The noun *top* functions as the subject of the gerund *gittikçe*, and this non-finite clause functions as an adverbial adjunct of the main verb.

- In (8d), the adjective *küçük* is used as a noun, and functions as the subject of the main verb. The noun phrase *kırmızı top* functions as the subject of the gerund *gittikçe*, and this non-finite clause functions as an adverbial adjunct of the main verb.

- In (8e), the noun phrase *küçük kırmızı top* functions as the subject of the gerund *gittikçe* (cf. (8b) where it functions as the subject of the main verb), and this non-finite clause functions as an adverbial adjunct of the main verb. Note that the subject of the main verb in this interpretation (i.e., *it*) is a covert one. Hence, it does not appear in the c-structure shown in Figure 7.

---

[12] We should however note that the times reported are exclusive of the time taken by the morphological processor, which with a 24,000 word root lexicon is rather slow and can process about 2–3 lexical forms per second. We have, however, ported our morphological analyzer to the XEROX TWOL system developed by Karttunen and Beesley [Karttunen and Beesley 1992] and this system can process about 500 forms a second. We intend to integrate this to our system soon.

[13] Word order freeness in Turkish allows various kinds of discontinuous constituents, e.g., an adverbial adjunct cutting in the middle of a compound noun.

[14] This example is not in any of the texts mentioned above. It is taken from the first author's thesis [Güngördü 1993].

[15] In fact, this sentence has a fifth interpretation due to the lexical ambiguity of the second word. In Turkish, *kırmız* is the name of a shining, red paint obtained from an insect with the same name. So, (8a) also means '*His little red paint sped up as the ball went.*' However, this is very unlikely to come to mind even for native speakers.

[16] The c-structures given here are simplified by removing some nodes introduced by certain intermediary rules to increase readability.

[17] In Turkish, any adjective can be used as a noun.

```
;**** ambiguity 1 ***
((SUBJ
   ((*AGR* 3SG) (*CASE* NOM)
        (*DEF* -)
        (*CAT* NP)
        (MODIFIED
            ((*CAT* NP)
                (MODIFIER
                    ((*CASE* NOM) (*AGR* 3SG)
                        (*LEX* "kIrmIzI")
                        (*CAT* ADJ)
                        (*R* "kIrmIzI")))
                (MODIFIED
                    ((*CAT* N) (*CASE* NOM)
                        (*AGR* 3SG)
                        (*LEX* "top")
                        (*R* "top")))
                (*AGR* 3SG)
                (*CASE* NOM)
                (*LEX* "top")
                (*DEF* -)))
        (MODIFIER
            ((*SUB* QUAL) (*CASE* NOM)
                (*AGR* 3SG)
                (*LEX* "kUCUk")))
        )
 )
 (VERB
   ((*TYPE* VERBAL) (*VOICE* ACT)
                    (*LEX* "hIzlandI")
                    (*CAT* V)
                    (*R* "hIzlan")
                    (*ASPECT* PAST)
                    (*AGR* 3SG)))
 (ADVCOMPLEMENTS
     ((*SUB* TEMP) (*LEX* "gittikCe")
                   (*CAT* ADVP)
                   (*CONV*
                   ((*WITH-SUFFIX* "dikce")
                       (*CAT* V)
                       (*R* "git")))))))
```

Figure 3: Output of the parser for the first the ambiguous interpretation of (8a) (i.e., (8b)).

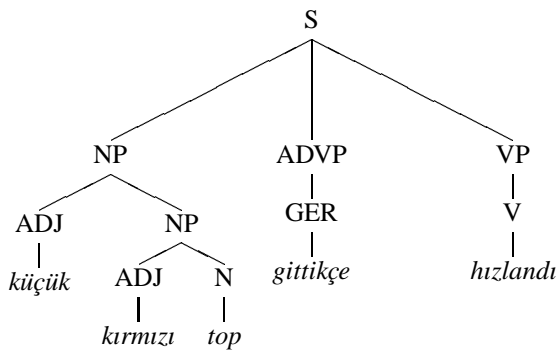

Figure 4: C-structure for (8b).

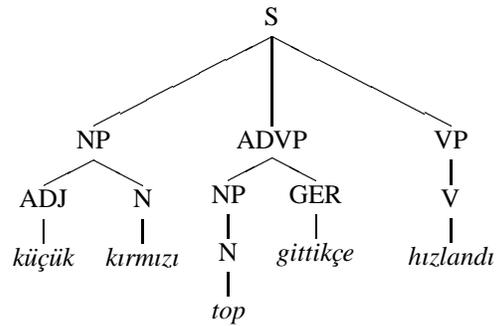

Figure 5: C-structure for (8c).

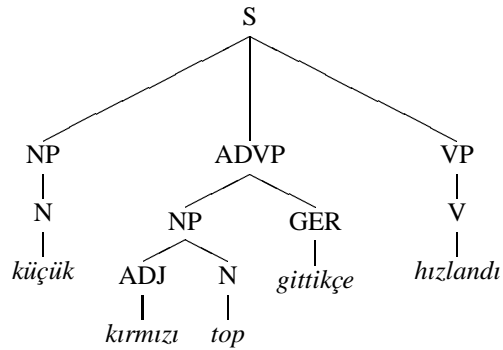

Figure 6: C-structure for (8d).

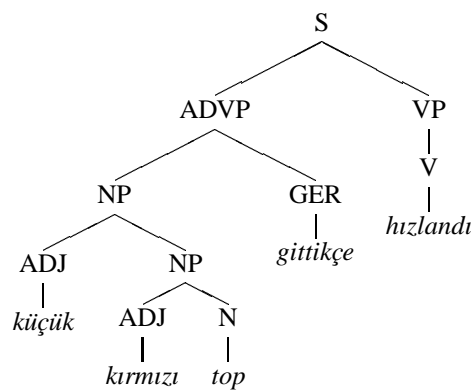

Figure 7: C-structure for (8e).

# 7  CONCLUSIONS AND SUGGESTIONS

We have presented a summary and highlights of our current work on providing an LFG specification for Turkish syntax. To the best of our knowledge this is the first such effort for constructing a computational grammar for Turkish. Our domain includes structurally simple and complex Turkish sentences. The rather complex morphological analyses of agglutinative words structures of Turkish are handled by a full-scale two-level morphological specification implemented in PC-KIMMO.

We have number of directions for improving our grammar and parser:

- Turkish is very rich in terms of adverbial constructs. We handle a great deal of these constructs by using a large number of rules. We are now in the process of developing a tagger with a multi-word construct recognizer to preprocess the text so that many multi-word and idiomatic constructs can be handled outside the grammar. In this way, multi-word constructs such as *yapar yapmaz* (do+AOR+3SG do+NEG+AOR+3SG) (*as soon as (one) does (that)*) where both lexical categories are verbal but the compound construct is an adverb, can be handled, so can idiomatic constructs like *yanı sıra* (side+3SG–POSS row) (*besides*) where the function and semantics of the multi-word construct has nothing to do with the function and semantics of the constituent lexical forms.

- We are currently working on extending the subset of sentences dealt with in respect of structure.

- We are currently working on augmenting our lexicon with substantial lexical information and selectional restriction information to be used with an integrated ontological database.

# 8  ACKNOWLEDGEMENTS

We would like to thank Carnegie-Mellon University, Center for Machine Translation for making available to us their LFG parsing system. We would also like to thank Elisabet Engdahl and Matt Crocker of the Centre for Cognitive Science, University of Edinburgh, for providing valuable comments and suggestions. This work was done as a part of a large scale NLP project (TU-LANGUAGE) which is funded by a NATO Grant under the Science for Stability Program.